\documentclass[a4paper, oneside]{article}

\usepackage{amsmath, amsthm, amssymb, graphicx, multicol, authblk, geometry, threeparttable, caption, subcaption, placeins, epstopdf, epsfig}

\usepackage[justification=centering]{caption}

\usepackage[table,xcdraw]{xcolor}

\title{Blockwise Compression of Transformer-based Models without Retraining}

\author{Gaochen Dong, Wei Chen}

\affil{Tensorchip, Beijing, China \\ gaochendong\_buaa@outlook.com}

\date{}

\begin{document}

\captionsetup[figure]{name={Fig.}}

\maketitle

{\centering\section*{abstract}}

Transformer-based models, exemplified by GPT-3, ChatGPT, and GPT-4, have recently garnered considerable attention in both academia and industry due to their promising performance in general language tasks. Nevertheless, these models typically involve computationally encoding processes, and in some cases, decoding processes as well, both of which are fundamentally large-scale matrix multiplication. These operations bring the inevitable challenges of massive computation resources and huge memory footprint, usually requiring at least $10^{23}$ FLOPs and hundreds of gigabytes, respectively. A common method to address this issue is to reduce the computational and memory requirements by applying layerwise quantization to the transformer, replacing the usual fp32 data type with a low-bit equivalent. Unfortunately, this method often leads to decreased model accuracy and necessitates time-consuming retraining. Such retraining not only requires fine-tuning skills but also substantial computational resources, posing challenges for users. To specifically tackle these issues, we propose BCT, a framework of blockwise compression for transformers without retraining, aiming to facilitate model deployment. Unlike layerwise compression methods, BCT achieves finer compression of the entire transformer by operating blockwise. This method mitigates data distribution deviation caused by quantization, eliminating the requirement for retraining. BCT effectively compresses all components of the model, including but not limited to the embedding, matrix multiplication, GELU, Softmax, layer normalization, and intermediate results. In a case study, an efficient model is compressed by BCT achieving up to 7.988x compression. Subsequently, we also evaluate it on several General Language Understanding Evaluation (GLUE) datasets. Experimental results on the majority of GLUE benchmark demonstrate the effectiveness of our method, as BCT achieves less than a 0.9\% degradation in accuracy compared to the more than a 1\% degradation seen with other methods providing similar or inferior compression ratios.

\noindent\textbf{\textit{Keywords:}}transformer; compression; blockwise; noretraining

\begin{multicols}{2}

\section{Introduction}

Transformer-based models\cite{vaswaniattention2017}, such as GPT-3\cite{brownlanguage2020}, ChatGPT and GPT-4, have demonstrated considerable promise and state-of-the-art performance in many fields, such as sentiment classification, machine translation, document analysis, question answering, text summarization, multi-round dialogue, image classification, visual question answering, and visual commonsense reasoning. As a result, the demand for deploying these models in business and scientific applications, such as advanced search engines, AI-powered chatbots for enhanced customer service, and scientific research support, has been rapidly increasing.

However, the widespread use of transformer-based models is hindered by their substantial computation and memory requirements, primarily due to the utilization of the multi-head self-attention mechanism. Usually, they require at least $10^{23}$ FLOPs and hundreds of gigabytes, respectively. High requirements for hardware platforms bring incredible difficulties to deployment and application.

To address these challenges, model compression has emerged as a feasible strategy to reduce the computational and memory requirements of transformer-based models while maintaining their predictive performance. Compressed models offer faster calculations, reduced memory footprint, and lower bandwidth requirements, thereby facilitating deployment and application. In this case, several compression methods are proposed to compress transformer-based models with an acceptable degradation of accuracy. However, most of them require extensive retraining, which can take weeks or even months, to fine-tune the parameters using a prepared-well calibration dataset to match the original data distribution. 

In this paper, we introduce Blockwise Compression of Transformers without retraining (BCT), a novel framework that compresses each component of the transformer. Unlike previous compression frameworks that may only focus on specific components, BCT utilize a blockwise method to compress the entire transformer, including the embedding, matrix multiplication, GELU\cite{hendrycksgaussian2020}, Softmax, layer normalization\cite{balayer2016}, and all the intermediate results. Fig.1 illustrates the framework of BCT. Experiments demonstrate that, without retraining, BCT can maintain an accuracy degradation of less than 0.9\% in most tasks on the GLUE\cite{wangglue2019} datasets.

\begin{figure*}[htbp]
\centering
\includegraphics[width=0.75\linewidth]{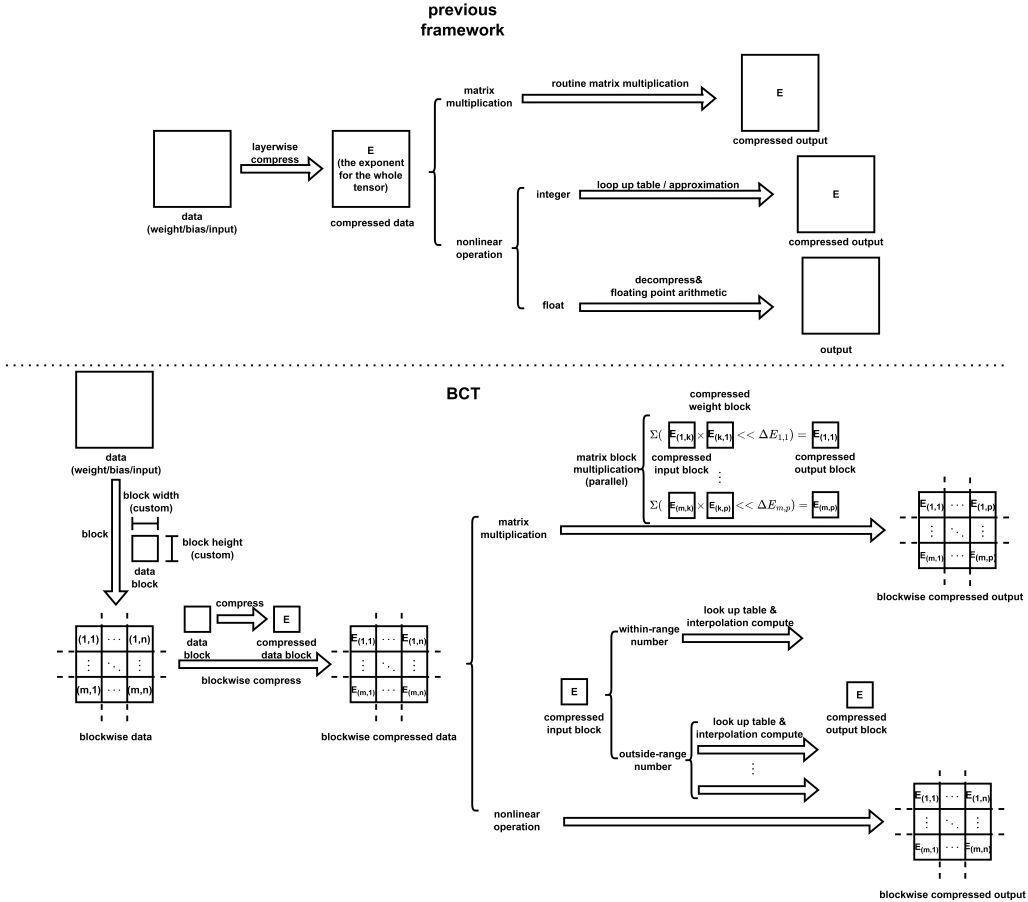}
\caption{The Framework of BCT}
\end{figure*}

This paper makes three major contributions: (1) We propose BCT, an efficient blockwise compression framework for transformer-based models that eliminates the need for retraining. (2) A practical and efficient methodology is proposed to handle each transformer layer, including matrix multiplication and nonlinear operations. (3) We evaluate BCT on multiple GLUE datasets, demonstrating its effectiveness with an accuracy degradation of less than 0.9\% on most tasks, surpassing many existing methods. Overall, BCT provides an efficient and practical solution for deploying transformer-based models in resource-constrained environments without retraining, significantly reducing the barriers to their wider application.

\section{Related Work}

In recent years, the compression of transformer-based models has emerged as a prominent area of research. Compressed models can significantly reduce computational resource requirements and memory footprints. Several efficient and practical compression methods, including pruning and quantization, have been proposed. For instance,  SparseGPT\cite{frantarsparsegpt2023} achieves 60\% sparsity by pruning GPT without necessitating retraining. Methods such as Q8BERT\cite{zafrirq8bert2019} and Q-BERT\cite{shenq-bert2019} implement partial quantization on BERT, while using floating-point arithmetic for the remainder, an method commonly referred to as 'fake quantization'. However, above methods fail to achieve the optimal compression ratio. I-BERT\cite{kimi-bert2021} quantizes BERT with integer-only arithmetic and infers matrix multiplication with int8 and nonlinear operations with int32. FQ-BERT\cite{liuhardware2021} also fully quantizes the whole BERT and can be deployed on FPGA. Both methods achieve a high compression ratio, especially 7.94x compression for FQ-BERT. However, they both need retraining to fine-tune the parameters. In contrast to these exsiting methods, our proposed framework, BCT, achieves a high compression ratio while eliminating the need for retraining. BCT can maintain an accuracy degradation of less than 0.9\% in most tasks with 7.988x compression, even higher than FQ-BERT.

\section{Method}
In this paper, BCT compresses the parameters and intermediate results of the model with blockwise compression and performs matrix multiplication and nonlinear operations with low-bit arithmetic, as shown in Fig.1.

\subsection{Blockwise Compression}

\begin{figure*}[htbp]
\centering
\includegraphics[width=0.6\linewidth]{block\_vs\_layer.pdf}
\caption{An Example of Blockwise and Layerwise Compression {\protect \\ \small Initialize a 4 * 4 matrix and set block size = 2. \protect \\ \small The number in the center of the dashed box is the shift value corresponding to the number in the dashed box}}
\end{figure*}

\begin{figure*}[htbp]
\centering
\includegraphics[width=0.4\linewidth]{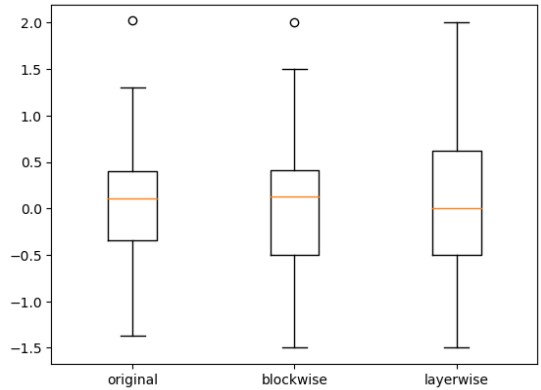}
\caption{Comparison of Data Distribution through Box Plot}
\end{figure*}

BCT partitions the weights, biases and intermediate results of a model into blocks and compresses them accordingly. The dimensions of these blocks are determined by the specific hardware platform in use. It is essential to strike a balance between hardware efficiency and model performance when selecting the appropriate block size. Generally, 64 is a block size that is friendly to both model parameter size and memory size. Unlike layerwise compression, which affects the data distribution on a larger scale, blockwise compression brings about changes at a smaller scale and is less influenced by the data distribution of other blocks. Therefore, the distribution of blockwise compressed data is closer to the original distribution. Fig.2 provides an in-depth comparative study between blockwise and layerwise compression algorithms. Fig.3, by employing a box plot representation, effectively elucidates the data distribution for the original data in conjunction with both compression methods in Fig.2. Consequently, the distribution of the blockwise compressed data closely resembles the original distribution. Additionally, blockwise compression takes into account the correlation information among samples, resulting in lower accuracy degradation further.

Due to the superior performance of blockwise compression, BCT can compress weight, bias and intermediate results to low-bit integer data or low-bit floating point data, thereby saving computation resources and reducing memory footprints. Fig.4 shows the memory footprint of various data types.

\begin{figure*}[htbp]
\centering
\includegraphics[width=0.6\linewidth]{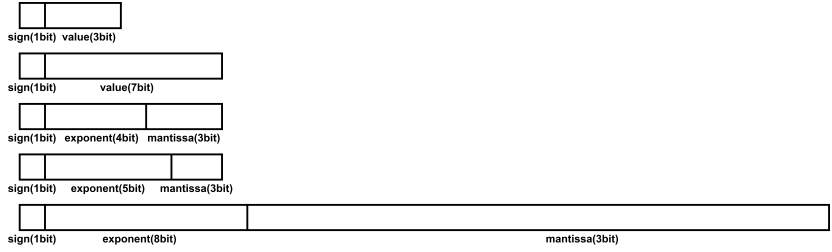}
\caption{The Bitwidth of Data Types}
\end{figure*}

\begin{table*}[btp]
\centering
\caption{The Representation of Data Types}
\begin{tabular}{lll}
\hline
Data Type & Dynamic Range                                               & Min Positive Value           \\ \hline
int4      & -8$\sim$7                                                   & 1                            \\
int8      & -128$\sim$127                                               & 1                            \\
fp8(e4m3) & -240$\sim$240                                               & 1.95*10\textasciicircum{}-3  \\
fp8(e5m2) & -57344$\sim$57344                                           & 1.526*10\textasciicircum{}-5 \\
fp32      & -3.4*10\textasciicircum{}38$\sim$3.4*10\textasciicircum{}38 & 1.4*10\textasciicircum{}-45  \\ \hline
\end{tabular}
\end{table*}

In this paper, we apply a symmetric shift quantization method\cite{miyashitaconvolutional2016} to compress fp32 data into low-bit integer data, which is more hardware-friendly than linear compression appiled by previous compression method. For k-bit compression of block x, the functions are:

\begin{equation}
shift = \lfloor log_{2}(\frac{2^{k-1}}{max(|x|)}) \rfloor 
\end{equation}

\begin{equation}
x_c = clip(\left[ x << shift \right], MIN, MAX)
\end{equation}

where c means compression and MIN, and MAX are calculated statistically on the calibration dataset by measuring KL divergence. This enables each compressed block, denoted as $x_c$, to have its own shift value, which can be seen as the exponent section of the block. In contrast, when compressing data layerwise, all the compressed data share a single exponent section, which means coarser-grained quantization and greater data distribution deviation. In addition to integer data, BCT can clip the exponent and mantissa sections of fp32 data to obtain fp8 data. The representation ranges of the data types we use are shown in Table 1.

\subsection{Matrix Multiplication}
The core concept lies in the observation that matrix multiplication can be performed as matrix block multiplication. For instance, let us consider the equation $ Y=X \cdot W^{T} + B $, where X represents the input, Y represents the output, W represents the weights, and B represents the biases. In BCT, X, W, and B are partitioned into blocks, which are then blockwise compressed to obtain cX, cW, and cB respectively. Subsequently, BCT conducts matrix block multiplication $cX\cdot cW^{T}$ and adds the compressed bias cB to yield the final output cY.

When performing matrix multiplication, there are certain crucial details to consider. Prior to accumulation or addition, it is necessary to normalize the exponent sections(the shift values) of the blocks. This normalization process is illustrated in Fig.5. Specifically, the exponent section $shift_{k}$ of each block $ cX_{ik}\cdot cW_{jk}^{T}$ should be adjusted uniformly to match the maximum exponent section $shift_{max}=max(shift_{k})$.

The overall process for each accumulation step can be outlined as follows:

\begin{equation}
acc_{ij} = \sum_{k=1}^{q}\left[ cX_{ik}\cdot cW_{jk}^{T} << (shift_{max} - shift_k) \right]
\end{equation}

In addition, the exponent sections of $acc_{ij}$ and $cC_{j}$ should be uniform before adding them.

\end{multicols}
\begin{figure}[htbp]
\centering
\includegraphics[width=0.6\linewidth]{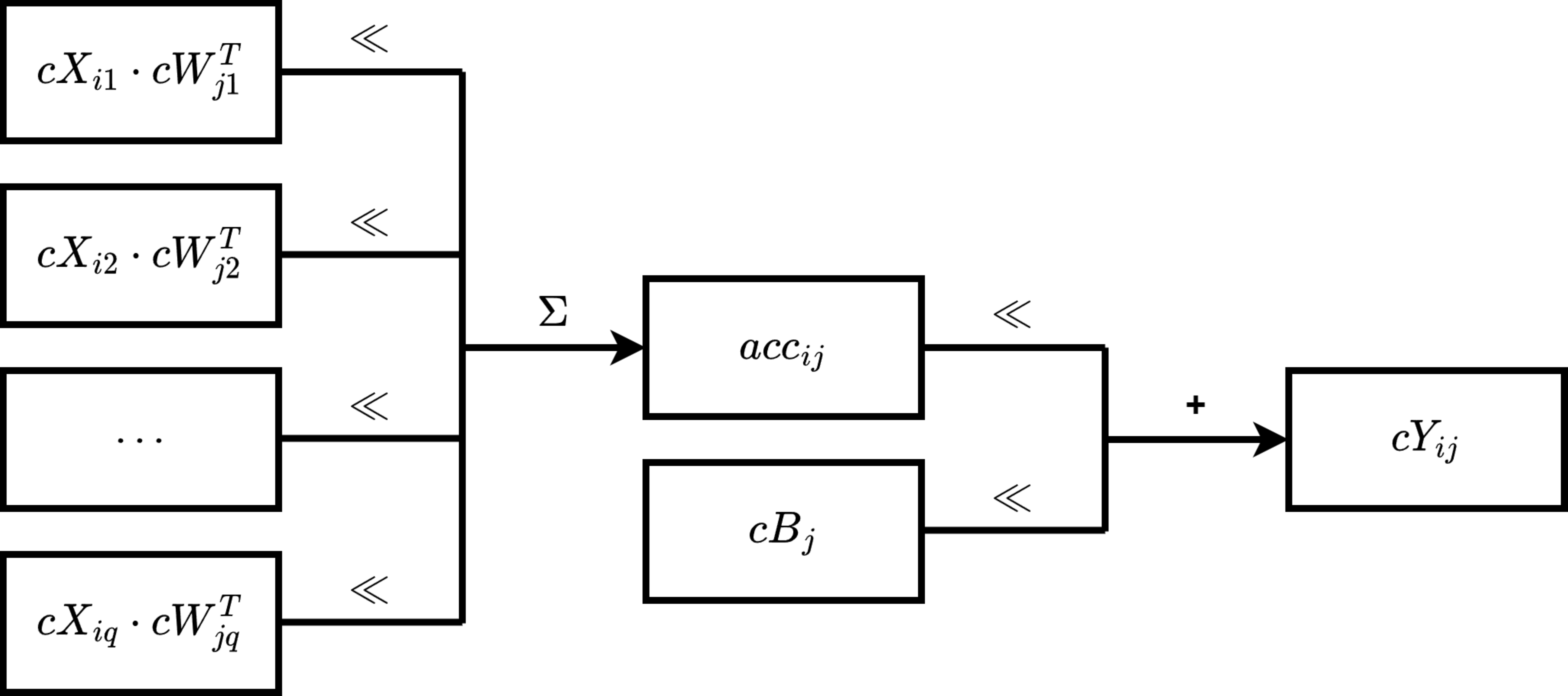}
\caption{Normalize Exponent Sention Before Accumulation or Addition}
\end{figure}
\begin{multicols}{2}

\begin{figure*}[htbp]
\centering
\includegraphics[width=0.8\linewidth]{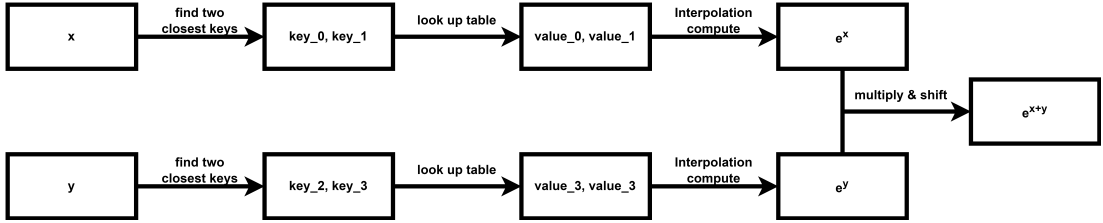}
\caption{Process Out-of-Range Data in Nonlinear Exponential Operation}
\end{figure*}

In addition to mitigating distribution deviation and consequently eliminating the need for retraining, implementing block-wise compression and computation in matrix multiplication operations proffers another significant advantage. Specifically, it enhances the parallelism inherent in matrix multiplications. This process involves customizing the block size and treating each block multiplication - involving the multiplication of an input data block $cX_{ik}$ with a weight block $cW_{jk}^{T}$ - as an independent entity. This independence allows for the simultaneous execution of block-wise matrix multiplication. As a result, computational resources can be utilized more efficiently, computational speed is increased, and overall performance is improved. Therefore, the block-wise method not only optimizes resource allocation but also boosts the cost-effectiveness and performance of transformer-based models.

\subsection{Nonlinear Operations} 
For nonlinear operations like GELU, Softmax, and LayerNorm, a pragmatic method involves computation with low-bit floating-point data such as fp8 or employing an int8 lookup table. In the case of the latter, we confine the input within a specific sampling range and treat the compressed input as keys. Concurrently, the output corresponding to each input is compressed and regarded as values. Consequently, we construct a lookup table consisting of 256 key-value pairs that can accommodate any int8 input. However, instead of directly obtaining the output from this table, we deploy an interpolation algorithm to approximate these nonlinear operations. This method results in a smaller degradation in accuracy. To illustrate, consider two keys, `$key_0$` and `$key_1$`, which are closest to the input x. Upon looking up the table, we receive corresponding values `$value_0$` and `$value_1$`. The output y is then calculated using a weighted average formula:

$$ y = \frac{x - key_1}{key_0 - key_1}value_0 + \frac{x - key_0}{key_1 - key_0}value_1 $$

This practical method enables a more accurate approximation of nonlinear operations and contributes to the enhanced efficiency of our proposed blockwise compression method.

\subsubsection{GELU}
The Gaussian Error Linear Unit (GELU), an activation function widely involved in the transformer-based models, is mathematically defined as follows:

\begin{equation}
GELU(x)=x\cdot \frac{1}{2} \left[ 1+erf(x/\sqrt{2}) \right]
\end{equation}

Owing to its superb linear properties, analogous to the Rectified Linear Unit (ReLU)\cite{xuempirical2015}, GELU becomes a suitable candidate for interpolation algorithms. These exceptional characteristics enable us to compress the input and output of GELU directly with minimal degradation in accuracy. Consequently, by constructing a lookup table and implementing the interpolation algorithm based on the compressed input and output, we can maintain a satisfactory level of accuracy while enhancing computational efficiency.
 
\subsubsection{Softmax}
The Softmax function, pivotal in transformer-based models, is mathematically represented as:

\begin{equation}
Softmax(x_i)=\frac{exp(x_i)}{\sum_{j}exp(x_j)}
\end{equation}

Central to the Softmax operation is a nonlinear exponential component. We address this by initiating the exponential operation via a table lookup method. Consequently, the output of Softmax can be derived through low-bit arithmetic operations. Prior to constructing a lookup table for the exponential operation, it proves beneficial to subtract all inputs by their respective maximum value. This action constrains the output of the exponential operation to within the range of 0 and 1, without altering the final output of the Softmax function. This strategy effectively circumvents the issue of an excessively broad output distribution from the exponential operation which could complicate compression. Furthermore, out-of-range data can be managed by leveraging the property $ e^{x+y} = e^{x}\times e^{y} $ of the exponential operation. By taking two inside-range numbers x and y as separate inputs, we can obtain the exponential value for the potentially outside-range sum, x+y, as depicted in Fig.6. This method allows us to aptly handle out-of-range data and maintain an effective blockwise compression method.

\subsubsection{LayerNorm}
The Layer Normalization (LayerNorm) function is a popular normalization method used in transformer-based models, which is defined as:

\begin{equation}
LayerNorm(x)=\frac{x-E[x]}{\sqrt{Var[x]+\varepsilon}}\times \gamma + \beta 
\end{equation}

At the core of the LayerNorm function lies a nonlinear component — the square root operation. In line with our methodology, we compress both the input and output of the square root function to construct a corresponding lookup table. Based on this table and employing the interpolation algorithm, we can derive the output of LayerNorm using low-bit arithmetic. Just as for the exponential operation, we devise a strategy to deal with data that falls outside the designated sampling range. We leverage the mathematical principle $\sqrt{x + y} = \sqrt{x} * \sqrt{y}$ to process such out-of-range data. Furthermore, the square root function also resides in the scale layer of the attention mechanism, where we apply the same aforementioned method. This consistent application across different components reinforces the efficacy and universality of BCT.

\subsection{Intermediate Results}
It is entirely feasible to partition intermediate results into blocks and apply compression accordingly. An imperative rule that governs this process stipulates that any transfer of intermediate results between two layers must occur using low-bit data types. In circumstances where the output from a layer does not meet this specification, it becomes necessary to reapply the compression strategy, thereby ensuring the output conforms to the low-bit data format. Consequently, the intermediate results can be used as input, being calculated in blocks with the blockwise parameters of the next layer. This method aligns with our broader methodology and contributes to maintaining robust computational efficiency across all layers.

\section{EXPERIMENTS}

\subsection{Dataset}

For an extensive evaluation of BCT, we select several datasets from GLUE benchmark. This includes one single-sentence task (SST2), one similarity and paraphrase task (STSB), and two inference tasks (MNLI and RTE). Such a diverse selection of tasks enables us to thoroughly scrutinize BCT's performance across a variety of application scenarios, thereby providing a comprehensive understanding of its capability, efficiency, and applicability in different scenarios.

\subsection{Setup}

Given that ChatGPT and GPT4 are not open-source, we have opted to employ the well-known BERT model as the testing ground for BCT. By the way, it is worth mentioning that other transformer-based models, such as ChatGPT and GPT4, employ similar operators to BERT in their model architectures. Therefore, the methodology of compressing BERT using BCT is equally applicable to these models. The baseline model used for comparisons is the BERT-base model furnished by Pytorch-Transformers\footnote{https://github.com/huggingface/transformers}. For a comprehensive analysis, we contrast BCT with two existing quantization methods - Q8BERT and FQBERT. Specifically, we devise four different models, each employing a unique data type for each layer, as detailed in Table 2. The BCT\_int8/fp32 model serves to validate the efficacy of blockwise compression without retraining, making it an apt comparison with Q8BERT. Notably, both BCT\_int8/fp32 and Q8BERT exclusively quantize the embedding and Feed-Forward Network (FFN) components of BERT. We design BCT\_int4/8 and BCT\_int8 to execute inference via low-bit integer arithmetic operations. On the other hand, our BCT\_fp8 model is configured to conduct inference utilizing low-bit floating-point arithmetic, demanding the least computational resources amongst the four models. It is closely trailed by BCT\_int4/8 in resource efficiency. Furthermore, we advocate for deploying the computation-heavy encoder/decoder elements of transformers in the accelerator. Concurrently, the embedding component, which demands negligible computational resources but has a significant impact on accuracy, can be computed in the CPU.

\end{multicols}

\begin{center}
\begin{table*}[htbp]\centering
\caption{Data Types of Four BCT Models}
\tabcolsep=0.3cm
\tiny
\begin{tabular}{llllllll}
\hline                                                                                     
models         & embedding(w/i) & Linear(w/b/i\tnote{*})  & MatMul(i)  & Softmax(i) & LayerNorm(w/b/i) & FFN(w/b/i)     & GELU(i) \\ \hline
BCT\_int8/fp32    & int8/int8 & fp32/fp32/fp32 & fp32   & fp32      & fp32/fp32/fp32    & int8/int8/int8       & fp32 \\ 
BCT\_int4/8   & fp32/fp32 & int4/int8/int8 & int8   & int8      & int8/int8/int8    & int4/int8/int8       & int8 \\
BCT\_int8 & fp32/fp32 & int8/int8/int8 & int8   & int8      & int8/int8/int8    & int8/int8/int8       & int8 \\
BCT\_fp8    & fp32/fp32 & fp8/fp8/fp8    & fp8      & fp8       & fp8/fp8/fp8       & fp8/fp8/fp8          & fp8  \\ \hline
\end{tabular}
\begin{tablenotes}
\footnotesize
\item[*] we denote w as the data type of weight, b as the data type of bias and i as the data type of intermediate results.
\end{tablenotes}
\end{table*} 
\end{center}

\begin{multicols}{2}

\subsection{Performance}

We use the accuracy of SST-2, MNLI, RTE, and the Spearman correlation of STS-B as comparison metrics. The benchmark results of GLUE tasks are shown in Table 3. 

\end{multicols}
\begin{center}
\begin{table*}[htbp]\centering
\caption{Benchmark Results of BCT}
\begin{tabular}{lllllllll}
\hline
            & SST-2   & STS-B   & MNLI-m   & RTE     \\ \hline
Bert-base   & 91.74\% & 87.36\% & 83.61\%  & 62.45\% \\
BCT\_int8/fp32 & 91.86\% & 87.39\% & 83.44\%        & 64.26\% \\
BCT\_int8 & 92.2\%  & 87.51\%   & 82.14\% & 63.81\% \\
BCT\_int4/8 & 90.94\%  & 86.49\%   & 80.08\% & 62.16\%       \\
BCT\_fp8 & 91.74\% & 87.36\% & 83.61\%        & 62.45\% \\ \hline
 \hline
\end{tabular}
\end{table*}
\end{center}

\begin{center}
\begin{table}[htbp]\centering
\caption{Accuracy Degradation of Models}
\begin{tabular}{lllllllll}
\hline
            & SST-2   & STS-B   & MNLI-m  & RTE     \\ \hline
Q8BERT      & -0.13\% & -0.65\% & -       & -1.32\% \\ 
BCT\_int8/fp32 & +0.12\% & +0.03\%       & -0.17\%       & +1.81\% \\ \hline
FQ-BERT     & -0.81\% & -       & -3.61\% & -       \\
BCT\_int8 & +0.46\% & +0.15\% & -1.47\% & +1.36\% \\
BCT\_int4/8 & -0.80\%       & -0.87\%       & -3.53\%  & -0.29\%       \\ \hline
BCT\_fp8 & +0.00\%  & +0.00\%  & +0.00\%  & +0.00\%  \\ \hline
\end{tabular}
\end{table}
\end{center}
\begin{multicols}{2}

In the majority of tasks, BCT\_int8/fp32 exhibits a lower decrease in accuracy compared to Q8BERT. This can be attributed to the blockwise compression strategy employed by BCT. Furthermore, BCT\_int4/8 demonstrates a higher level of compression(7.988x) when compared to FQ-BERT(7.94x). For instance, while FQ-BERT employs a 32-bit bias, BCT\_int4/8 utilizes an 8-bit bias. Additionally, it is worth noting that BCT\_int4/8 showcases a mere 0.80\% degradation in accuracy on the SST-2 dataset, which is even less than the 0.81\% reduction observed in FQ-BERT, all without requiring retraining. This substantiates the robustness and effectiveness of BCT\_int4/8. Moreover, BCT\_int8 and BCT\_fp8 deliver outstanding performance across various datasets. Through these advancements, BCT-based models have exhibited a superior capability to compress models with minimal degradation in accuracy. 

\section{Conclusion}

In this study, we introduced BCT that efficiently blockwise compresses transformer-based models without retraining. Unlike traditional compression methods that operate at the layer level, BCT works at a finer block level across the entire model. BCT's compression using fp8 or int4 data types achieves a balance between performance and resource consumption. It applies uniformly across all components of the transformer model, effectively eliminating data distribution deviation caused by quantization and negating the need for retraining. Furthermore, BCT’s hardware-friendly shift compression method enhances computation efficiency. Practical benefits of BCT are evident in deploying transformer-based models. A case study showed BCT could compress a model to 7.988x of its original size, maintaining competitive performance. Testing on various GLUE datasets also demonstrated BCT's effectiveness, with less than a 0.9\% degradation if accuracy in most tasks compared to alternative methods. Our work confirms that BCT substantially lowers computational and memory demands while preserving near-original model performance levels. In conclusion, BCT provides a practical solution for the deployment of transformer-based models, especially Language Learning Models (LLMs) like ChatGPT and GPT4, overcoming technical and economic challenges in numerous scenarios.

\bibliographystyle{apalike}

\bibliography{BCT}

\begin{thebibliography}{}

\bibitem[Ba et~al., 2016]{balayer2016}
Ba, J.~L., Kiros, J.~R., and Hinton, G.~E. (2016).
\newblock Layer normalization.
\newblock {\em arXiv preprint arXiv:1607.06450}.

\bibitem[Brown et~al., 2020]{brownlanguage2020}
Brown, T., Mann, B., Ryder, N., Subbiah, M., Kaplan, J.~D., Dhariwal, P.,
  Neelakantan, A., Shyam, P., Sastry, G., Askell, A., et~al. (2020).
\newblock Language models are few-shot learners.
\newblock {\em Advances in neural information processing systems},
  33:1877--1901.

\bibitem[Frantar and Alistarh, 2023]{frantarsparsegpt2023}
Frantar, E. and Alistarh, D. (2023).
\newblock Massive language models can be accurately pruned in one-shot.
\newblock {\em arXiv preprint arXiv:2301.00774}.

\bibitem[Hendrycks and Gimpel, 2016]{hendrycksgaussian2020}
Hendrycks, D. and Gimpel, K. (2016).
\newblock Gaussian error linear units (gelus).
\newblock {\em arXiv preprint arXiv:1606.08415}.

\bibitem[Kim et~al., 2021]{kimi-bert2021}
Kim, S., Gholami, A., Yao, Z., Mahoney, M.~W., and Keutzer, K. (2021).
\newblock I-bert: Integer-only bert quantization.
\newblock In {\em International conference on machine learning}, pages
  5506--5518. PMLR.

\bibitem[Liu et~al., 2021]{liuhardware2021}
Liu, Z., Li, G., and Cheng, J. (2021).
\newblock Hardware acceleration of fully quantized bert for efficient natural
  language processing.
\newblock In {\em 2021 Design, Automation \& Test in Europe Conference \&
  Exhibition (DATE)}, pages 513--516. IEEE.

\bibitem[Miyashita et~al., 2016]{miyashitaconvolutional2016}
Miyashita, D., Lee, E.~H., and Murmann, B. (2016).
\newblock Convolutional neural networks using logarithmic data representation.
\newblock {\em arXiv preprint arXiv:1603.01025}.

\bibitem[Shen et~al., 2020]{shenq-bert2019}
Shen, S., Dong, Z., Ye, J., Ma, L., Yao, Z., Gholami, A., Mahoney, M.~W., and
  Keutzer, K. (2020).
\newblock Q-bert: Hessian based ultra low precision quantization of bert.
\newblock In {\em Proceedings of the AAAI Conference on Artificial
  Intelligence}, volume~34, pages 8815--8821.

\bibitem[Vaswani et~al., 2017]{vaswaniattention2017}
Vaswani, A., Shazeer, N., Parmar, N., Uszkoreit, J., Jones, L., Gomez, A.~N.,
  Kaiser, {\L}., and Polosukhin, I. (2017).
\newblock Attention is all you need.
\newblock {\em Advances in neural information processing systems}, 30.

\bibitem[Wang et~al., 2018]{wangglue2019}
Wang, A., Singh, A., Michael, J., Hill, F., Levy, O., and Bowman, S.~R. (2018).
\newblock Glue: A multi-task benchmark and analysis platform for natural
  language understanding.
\newblock {\em arXiv preprint arXiv:1804.07461}.

\bibitem[Xu et~al., 2015]{xuempirical2015}
Xu, B., Wang, N., Chen, T., and Li, M. (2015).
\newblock Empirical {Evaluation} of {Rectified} {Activations} in
  {Convolutional} {Network}.
\newblock arXiv:1505.00853 [cs, stat].

\bibitem[Zafrir et~al., 2019]{zafrirq8bert2019}
Zafrir, O., Boudoukh, G., Izsak, P., and Wasserblat, M. (2019).
\newblock Q8bert: Quantized 8bit bert.
\newblock In {\em 2019 Fifth Workshop on Energy Efficient Machine Learning and
  Cognitive Computing-NeurIPS Edition (EMC2-NIPS)}, pages 36--39. IEEE.

\end{thebibliography}

\end{multicols}

\end{document}